# Exploratory State Representation Learning

*Astrid Merckling*, Nicolas Perrin-Gilbert, Alex Coninx and Stéphane Doncieux*

*Sorbonne Université, CNRS, Institut des Systèmes Intelligents et de Robotique, ISIR, Paris, France*

Not having access to compact and meaningful representations is known to significantly increase the complexity of reinforcement learning (RL). For this reason, it can be useful to perform state representation learning (SRL) before tackling RL tasks. However, obtaining a good state representation can only be done if a large diversity of transitions is observed, which can require a difficult exploration, especially if the environment is initially reward-free. To solve the problems of exploration and SRL in parallel, we propose a new approach called XSRL (eXploratory State Representation Learning). On one hand, it jointly learns compact state representations and a state transition estimator which is used to remove unexploitable information from the representations. On the other hand, it continuously trains an inverse model, and adds to the prediction error of this model a $k$-step learning progress bonus to form the maximization objective of a discovery policy. This results in a policy that seeks complex transitions from which the trained models can effectively learn. Our experimental results show that the approach leads to efficient exploration in challenging environments with image observations, and to state representations that significantly accelerate learning in RL tasks.

****Correspondence:***
*Astrid Merckling*
*astrid.merckling@isir.upmc.fr*

## 1 INTRODUCTION

Recent improvements in computational power and deep learning techniques have been combined with reinforcement learning (RL) to create deep RL (DRL) algorithms capable of solving complex control tasks with continuous state and action spaces (Li, 2018). These improvements have popularized end-to-end DRL techniques, which involve letting deep learning systems automatically learn representations and make predictions simultaneously (i.e. without performing a feature extraction as a preliminary phase). However, despite its simplicity of design, this end-to-end strategy has limitations [see Glasmachers (2017)] such as potential instability and slow convergence. In some cases it seems advantageous to separate representations and policies in different modules and train representations independently from the sparse and delayed rewards of RL tasks.

State-of-the-art end-to-end DRL algorithms face a significant computational challenge, especially in the context of continuous control tasks with visual observations (Kostrikov et al., 2020; Laskin et al., 2020). Instead of addressing this challenge directly, this paper focuses on the state representation learning (SRL) alternative. SRL focuses on solving the state representation learning problem independently of a control task, in order to make the inputs more machine-readable for DRL algorithms (Lange and Riedmiller, 2010; Jonschkowski and Brock, 2013; Böhmer et al., 2015). It relies on task-agnostic and reward-free interactions to capture relevant information about the agent and its environment and to represent it in a compact form (Lesort et al., 2018; Morik et al., 2019).

The main starting point of our work is the following remark: for state representations to be useful as inputs to new RL tasks, the SRL training must have observed a large diversity of transitions. In the SRL literature, this has typically been addressed with demonstrations (Sermanet et al., 2018; Merckling et al., 2020) or random exploration (Jonschkowski and Brock, 2015; Yarats et al., 2019). However, it is often impossible to randomly explore all environment transitions, and generating demonstrations requires time and a priori knowledge about potential tasks. Therefore, this work proposes to extend the exploration strategies used in RL to the context of SRL. We place ourselves in a pure exploration context, where no extrinsic reward is provided by the environment. A common approach with RL in this reward-free setting is to compute intrinsic rewards that estimate a degree of uncertainty about trained models (Bubeck et al., 2009; Shyam et al., 2019; Sekar et al., 2020). With this approach in mind, we propose a new exploration strategy to learn state representation models, called XSRL (eXploratory State Representation Learning).

XSRL consists of a twofold training procedure. In the first training procedure, XSRL learns state representations whose transitions are Markovian while advantageously reducing dimensionality by filtering out unexploitable information with respect to the objective of next observation prediction. In the second training procedure, XSRL learns discovery policies that perform actions considered uncertain by an inverse model. Finally, in order to cope with the two sources of non-stationarity due to changing state representations and inverse model predictions, we train two discovery policies in parallel and, given their performances, reset one of them after a given number of training steps (as explained in **Section 3.2.1**), where one training step corresponds to a gradient descent for some loss on a batch of transitions. We use an online training with a set of agents, each half of which follows one of the two policies (see **Section 3.3**).

The main contributions of XSRL can be summarized as follows. First, we introduce a novel SRL architecture based on recursive state estimation predictions. Second, XSRL provides an exploration strategy by optimizing discovery policies driven towards uncertain transitions (**Section 3.2**). Third, we demonstrate the validity of XSRL representations as well as its discovery policies through quantitative and qualitative evaluations on three different environments (**Section 5.1**). Finally, we show improvements over other representation strategies through a comparative quantitative evaluation on unseen control tasks with the popular RL algorithm SAC (Haarnoja et al., 2018) (**Section 5.2**).

## 2 RELATED WORK

Several other SRL algorithms with a near-future prediction objective have been proposed recently (Assael et al., 2015; Böhmer et al., 2015; Wahlström et al., 2015; Watter et al., 2015; van Hoof et al., 2016; Jaderberg et al., 2017; Shelhamer et al., 2017; de Bruin et al., 2018). However, they separately learn state representations from which current observations can be reconstructed, and train a forward model on the learned states. The main limitation of these approaches is the inefficiency of the reconstruction objective, which leads to representations that contain unnecessary information about the observations. Instead, XSRL jointly learns a state transition estimator with the next observation prediction objective. On the one hand, this forces the learned state representations to retrieve information and memorize it through the recursive loop in order to restore the observability of the environment (in this work, the partial observability is due to image observations) and to verify the Markovian property. On the other hand, this forces the learned state representations to filter out unnecessary information, in particular information about distractors (i.e. elements that are not controllable or do not affect an agent).

The XSRL exploration strategy is inspired by the line of work that maximizes intrinsic rewards corresponding to prediction errors of a trained forward model, which is a form of dynamics-based curiosity (Hester and Stone, 2012; Pathak et al., 2017; Burda et al., 2018). These strategies often combine intrinsic rewards with extrinsic rewards to solve the complex exploration/exploitation tradeoff. Instead, the first phase of XSRL ignores extrinsic reward to focus on SRL and prediction model learning. Extrinsic reward only comes in a second step (the RL tasks). In addition, for intrinsic motivation XSRL relies on prediction errors of an inverse model instead of those of a forward model. Prediction errors of an inverse model have the advantage of depending only on elements of the environment controllable by an agent (assuming there are no surjective transitions). It allows to discard the rest and thus to significantly reduces the size of the acquired state representation.

Finally, a variant of $k$-step learning progress bonus is used to focus on transitions for which the forward model predictions are changing. Learning progress estimation was initially proposed in the field of developmental robotics (Oudeyer et al., 2007). Lopes et al. (2012) initiated the estimation of learning progress bonuses to solve the exploitation/exploration tradeoff in the model-based RL domain with finite MDPs. Achiam and Sastry (2017) have scaled this approach to continuous MDPs with compact observations of several dozen dimensions. We apply the approach of Achiam and Sastry (2017) to image observations and in the SRL context.

## 3 PROPOSED METHOD: XSRL

### 3.1 State Transition Estimator

The goal of SRL is to transform high-dimensional observations into machine-readable compact representations which retrieve information about an agent and the environment (Lesort et al., 2018). With XSRL, we make the assumption that a good state representation must contain the information needed to predict the next observation from the previous time step, or at least the change in observation that can be explained by the agent's action.

Our state transition estimator $\varphi$ consists of two neural network parts ($\alpha$, $\beta$), and a common network head $\gamma$. While $\alpha$ is a convolutional neural network (CNN) to process image observations, $\beta$ is a multilayer perceptron (MLP) to process the concatenated action and state vectors. Finally, the common network head $\gamma$ is a MLP that processes the concatenated output vectors of the two first networks to estimate next state vectors ($\mathbf{s}_{t+1}$).

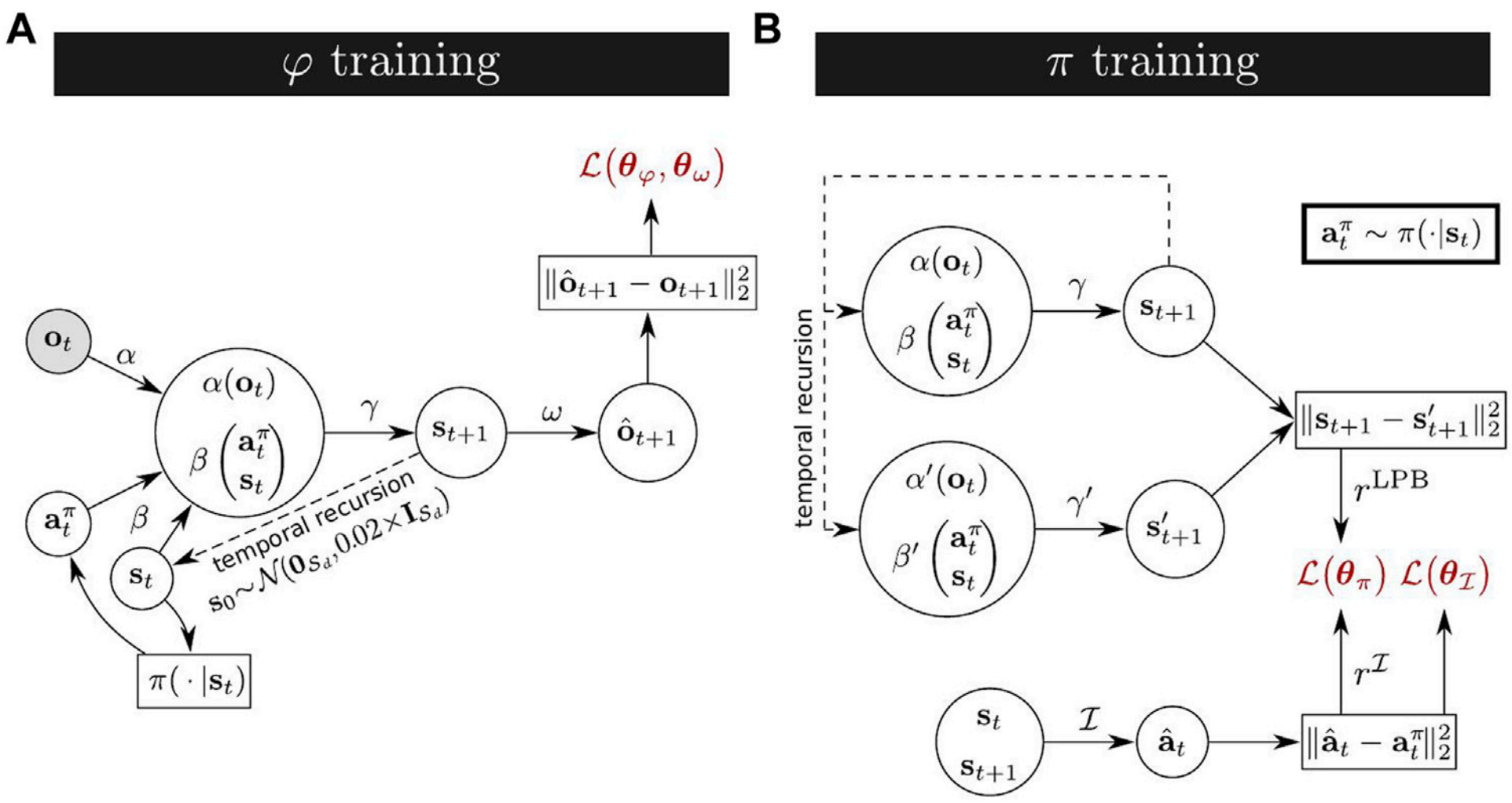

**FIGURE 1 | (A)** XSRL learning process of state representations by jointly training a state transition estimator $\varphi$ formed by ($\alpha$, $\beta$, $\gamma$) and a next observation predictor $\omega$; the action $\mathbf{a}_t^\pi$ is sampled from $\pi \in \{\pi_1, \pi_2\}$ with equal probability (a set of agents are considered in parallel and each is assigned a randomly chosen policy, as explained in **Section 3.3**). **(B)** XSRL learning process of a discovery policy by minimizing $\mathcal{L}(\theta_\pi)$, which is related to intrinsic rewards. Intrinsic rewards are formed of two main terms: (i) $r^\mathcal{I}$: prediction errors of an inverse model $\mathcal{I}$ (also used in $\mathcal{L}(\theta_\mathcal{I})$); (ii) $r^{\text{LPB}}$: $k$-step learning progress bonuses for $\varphi$ where the parameters of $\varphi'$ formed by ($\alpha'$, $\beta'$, $\gamma'$) are delayed by $k$ training steps and kept fixed.

The graph in **Figure 1A**, shows how, from current observation $\mathbf{o}_t$, action $\mathbf{a}_t$ and state $\mathbf{s}_t$, information is compactly merged into a next state $\mathbf{s}_{t+1}$ through the intermediate functions ($\alpha$, $\beta$, $\gamma$). Because of the recursive loop on the state representation, $\varphi$ bootstraps from an initial state drawn from a Gaussian distribution of mean zero and standard deviation 0.02. Putting all the functions together, we get the following definition of next state representation predictions:

$$\mathbf{s}_{t+1} = \varphi([\mathbf{o}_t, \mathbf{s}_t, \mathbf{a}_t]) = \gamma([\alpha(\mathbf{o}_t),\ \beta([\mathbf{s}_t, \mathbf{a}_t])]) \tag{1}$$

where we abbreviate the state transition estimator network ($\alpha$, $\beta$, $\gamma$) by $\varphi$ and their parameters into the following parameter set $\boldsymbol{\theta}_\varphi = \{\boldsymbol{\theta}_\alpha, \boldsymbol{\theta}_\beta, \boldsymbol{\theta}_\gamma\}$. The implementation details of the whole neural network are displayed in **Table 3**.

$\varphi$ is trained jointly with a next observation predictor $\omega$. $\omega$ is a CNN with transposed convolution layers[1] trained to deterministically predict from the outputs of $\varphi$ (i.e. $\mathbf{s}_{t+1}$) the next observations: $\omega(\mathbf{s}_{t+1}) = \hat{\mathbf{o}}_{t+1}$. This yields the following prediction error:

$$\|\hat{\mathbf{o}}_{t+1} - \mathbf{o}_{t+1}\|_2^2 \tag{2}$$

All the parameters of $\omega$ are gathered in a single parameter set $\boldsymbol{\theta}_\omega$. The corresponding training process is described with the complete XSRL training process in **Section 3.3**.

Thanks to this joint training of $\varphi$ and $\omega$, XSRL builds compact state representations which contain the information needed to predict the action consequences in the next observation. We assume that a ground truth state space exists that follows Markovian transitions. It is unknown and only image observations are available, making the environment partially observable, which may be due to perceptual aliasing or to the dynamics of the system that cannot be fully captured by an image. We therefore force $\varphi$ to memorize in the state representations (through the recursive loop) the information of past time steps in order to build a state space with Markovian transitions. Indeed, to predict the (predictable part of the) next observation with $\omega$, the next state representation $\mathbf{s}_{t+1} = \varphi([\mathbf{o}_t, \mathbf{s}_t, \mathbf{a}_t])$ must contain the information of past and current time steps. As this information cannot only be retrieved from $\mathbf{o}_t$ and $\mathbf{a}_t$, some of it must be memorized in $\mathbf{s}_t$ through the recursive state loop. In this way, the state representations learned by XSRL are trained to form Markovian transitions that translate mathematically as follows:

$$P(\mathbf{s}_{t+1}|\mathbf{s}_t, \mathbf{a}_t) = P(\mathbf{s}_{t+1}|\mathbf{s}_t, \mathbf{a}_t, \mathbf{s}_{t-1}, \mathbf{a}_{t-1}, \ldots, \mathbf{s}_0, \mathbf{a}_0) \tag{3}$$

for all states $\mathbf{s}_{t+1}, \mathbf{s}_t \in \mathcal{S} \subset \mathbb{R}^{\mathcal{S}_d}$ and actions $\mathbf{a}_t \in \mathcal{A} \subset \mathbb{R}^{\mathcal{A}_d}$.

As perceptual aliasing may occur, $\varphi$ needs to encode information about previous steps to predict the right observations after ambiguous ones. For example, in the case of a mobile robotics setup (such as the TurtleBot Maze environment described below), the representation built by XSRL is expected to capture the topology of the environment because a form of odometry is necessary to predict next observations [see Böhmer et al. (2013)].

## 3.2 Discovery in the Face of Uncertainty

### 3.2.1 Over-Commitment

A problem that arises in pure exploration with dynamics-based curiosity is the non-stationarity of intrinsic rewards. Specifically,

[1]We used the 2D transposed convolution operator provided by PyTorch.

as in other dynamics-based curiosity explorations from image observations, two sources of non-stationarity emerge (Burda et al., 2018): (i) the models change and adapts to novel observations, which modifies intrinsic rewards, (ii) the state representations change, which requires further adaptation of the models. Such a non-stationary training signal can lead to slow exploration as policies have to "unlearn" in areas where the novelty wears off. Shyam et al. (2019) have called this problem "over-commitment." They proposed to circumvent it by training from scratch a new policy. We follow a similar idea in XSRL by training two discovery policies in parallel called $\pi_1$ and $\pi_2$, and every $T_{\text{reset}}$ iterations we reset the policy with the lowest cumulative intrinsic reward.

### 3.2.2 Intrinsic Rewards

The intrinsic rewards to be maximized by XSRL discovery policies are a combination of the following terms: (i) prediction errors of an inverse model which should be maximized on transitions with high uncertainty with respect to the elements controllable by an agent; (ii) $k$-step learning progress bonuses that should be maximized on transitions for which the predictions of the forward model $\varphi$ are changing; (iii) a policy entropy estimation to improve convergence stability. **Figure 1B** shows the graph corresponding to the calculation of the two main terms (i) and (ii).

#### *3.2.2.1 Inverse Model*

Previous dynamics-based curiosity methods typically used a forward model to indirectly estimate action uncertainty (Burda et al., 2018). The common issue with this approach is that it can drive exploration policies towards transitions with intrinsic (aleatoric) uncertainty (Schmidhuber, 1991). One way to solve this problem is to train an ensemble of models (Chua et al., 2018). Initialized differently, the models tend to disagree in neighborhoods of transitions that have not been explored (so where there is a lack of data, i.e. epistemic uncertainty), but the models agree on transitions that have been observed, even if they contain irreducible aleatoric uncertainty. Thus, seeking transitions for which the models disagree drives the exploration towards epistemic uncertainty, which is the desired behavior. In this paper, we follow another approach, inspired by Pathak et al. (2017), that combines a forward and an inverse model. In Pathak et al. (2017), the inverse model is used to construct a feature space that erases environmental features that are not influenced by the agent's actions. Curiosity based on a forward model in this feature space avoids the issue of aleatoric uncertainty. We proceed in a slightly different way that removes the need for a new feature space, by encouraging discovery policies to seek transitions for which the composition of a forward model ($\varphi$) and inverse model does not retrieve the intended action. This tends to be true when data is lacking (epistemic uncertainty), and false for data on which the models are well-trained. Aleatoric uncertainty is ignored by the forward model, but that does not prevent the inverse model from retrieving the correct action, therefore aleatoric uncertainty alone does not attract exploration.

Our inverse model takes as input a pair of consecutive states estimated by the forward model ($\mathbf{s}_t$, $\mathbf{s}_{t+1}$) to predict the action $\hat{\mathbf{a}}_t = \mathcal{I}(\mathbf{s}_{t+1}, \mathbf{s}_t)$ executed by an agent to obtain the next state $\mathbf{s}_{t+1}$. The prediction errors to be maximized by the discovery policies and minimized by the inverse model are calculated as follows:

$$r^{\mathcal{I}}(\hat{\mathbf{a}}_t, \mathbf{a}_t^\pi) = \left\|\hat{\mathbf{a}}_t - \mathbf{a}_t^\pi\right\|_2^2. \quad (4)$$

The action $\mathbf{a}_t^\pi$ is sampled from $\pi \in \{\pi_1, \pi_2\}$ with equal probability. The training process of the inverse model is detailed later in **Section 3.3**.

#### *3.2.2.2 Learning Progress Bonus*

To ensure that actions considered uncertain by the composition of a forward and inverse model lead to diverse unknown transitions, we use a $k$-step learning progress bonus on $\varphi$. It makes the agent curious mainly about things that change the predictions of $\varphi$. Following Achiam and Sastry (2017), we compute this learning progress bonus from $\varphi$ and its clone denoted $\varphi'$ formed by ($\alpha'$, $\beta'$, $\gamma'$), whose parameters are delayed by $k$ training steps and kept frozen. The squared Euclidean distance between the outputs of these two networks is an estimate of the changes in $\varphi$ after $k$ training steps.

The $k$-step learning progress bonus to be maximized by the two discovery policies is defined as follows:

$$r^{\text{LPB}}(\mathbf{o}_t, \mathbf{s}_t, \mathbf{a}_t^\pi) = \left\|\varphi([\mathbf{o}_t, \mathbf{s}_t, \mathbf{a}_t^\pi]) - \varphi'([\mathbf{o}_t, \mathbf{s}_t, \mathbf{a}_t^\pi])\right\|_2^2 \quad (5)$$

where the action $\mathbf{a}_t^\pi$ is sampled from $\pi \in \{\pi_1, \pi_2\}$ with equal probability.

#### *3.2.2.3 Policy Entropy Estimation*

Ziebart et al. (2008) and Haarnoja et al. (2017) showed that optimizing policies to maximize entropy in addition to expected return improved their convergence. The formulation depends on a temperature $w_{\mathcal{H}}$ which is the weight of the entropy maximization term. Following Haarnoja et al. (2018), the temperature tuning is automated by formulating a different entropy objective, where the entropy is treated as a constraint. Approximating a dual gradient descent, $w_{\mathcal{H}}$ is adapted online by gradient steps on the following expression:

$$w_{\mathcal{H}}[\mathcal{H}(\pi(\cdot|\mathbf{s}_t)) - \bar{\mathcal{H}}] \quad (6)$$

By default, $\bar{\mathcal{H}}$, the target entropy, is chosen to be equal to minus the action dimension $-\mathcal{A}_d$. See (Haarnoja et al., 2018) for more details.

### 3.2.3 Discovery Policies

Now that we have detailed the three terms for computing intrinsic rewards, we explain how we train discovery policies to maximize them. In this work, we study environments with continuous action spaces. A possible approach to learn a policy in this case is to model it as a multivariate Gaussian distribution with a diagonal covariance matrix (Haarnoja et al., 2018). To do this, we use a neural network with a first common part, then one head $\mu_\pi$ with parameters $\boldsymbol{\theta}_\mu$ to predict a mean vector, and a second head $\Sigma_\pi$ with parameters $\boldsymbol{\theta}_\Sigma$ to predict the diagonal covariance elements of a covariance matrix. The outputs of these two heads, which have the same dimension as the action space, allow us to parameterize a policy, so that it follows a Gaussian distribution defined as:

$$\pi(\cdot|\mathbf{s}_t) \triangleq \mathcal{N}\left(\mu_\pi(\mathbf{s}_t), \Sigma_\pi(\mathbf{s}_t)\right) \tag{7}$$

All parameters of a discovery policy $\pi$ are gathered in a single parameter set $\boldsymbol{\theta}_\pi = \{\boldsymbol{\theta}_\mu, \boldsymbol{\theta}_\Sigma\}$. The reparametrization trick (Kingma and Welling, 2014) is used to sample an action from a policy (i.e. $\mathbf{a}_t^\pi \sim \pi(\cdot|\mathbf{s}_t)$) to keep all its parameters differentiable:

$$\mathbf{a}_t^\pi \triangleq \mu_\pi(\mathbf{s}_t) + \boldsymbol{\epsilon}_t \times \Sigma_\pi(\mathbf{s}_t) \quad , \quad \boldsymbol{\epsilon}_t \sim \mathcal{N}\left(\mathbf{0}_{\mathcal{A}_d}, \mathbf{I}_{\mathcal{A}_d}\right) \tag{8}$$

The two discovery policies ($\pi \in \{\pi_1, \pi_2\}$) can be optimized directly from the intrinsic reward gradients. The intrinsic rewards are computed with prediction errors of an inverse model, $k$-step learning progress bonuses on $\varphi$, and a policy entropy estimation, all of which use actions sampled from $\pi$. Thus, our discovery policy training strategy is based on stochastic gradients from batches for the minimization of the expected value of the following loss function:

$$-\left(w_{\mathcal{I}} r^{\mathcal{I}}\left(\hat{\mathbf{a}}_t, \mathbf{a}_t^\pi\right) + w_{\text{LPB}} r^{\text{LPB}}\left(\mathbf{o}_t, \mathbf{s}_t, \mathbf{a}_t^\pi\right) + w_{\mathcal{H}} \mathcal{H}\left(\pi(\cdot|\mathbf{s}_t)\right)\right). \tag{9}$$

This yields a maximization of the intrinsic rewards. The corresponding training process is described in the next section.

## 3.3 Optimization Process

Let us define the notations for the training examples we manipulate in our online training procedure. There is an even number $B \geq 2$ of agents in parallel, denoted by $b \in [\![1, B]\!]$, each of them being initialized in the same fixed configuration. At time step $t$, a training example for $(\varphi, \omega)$ is an element of the form $(\mathbf{o}_{t+1}^{(b)}, \mathbf{o}_t^{(b)}, \mathbf{s}_t^{(b)}, \mathbf{a}_t^\pi(b))$, composed respectively of next observation, current observation, previously estimated state representation, and executed action sampled from one of the two discovery policies as $\mathbf{a}_t^\pi(b) \sim \pi(\cdot|\mathbf{s}_t^{(b)})$ (following the sampling process defined in **Eq. 8**). Specifically, each half of the set of $B$ agents follows one of the two policies $\pi \in \{\pi_1, \pi_2\}$. A state transition estimator $\varphi$ composed of three modules $(\alpha, \beta, \gamma)$ estimates from the triplet input $(\mathbf{o}_t^{(b)}, \mathbf{s}_t^{(b)}, \mathbf{a}_t^\pi(b))$ the next state $\mathbf{s}_{t+1}^{(b)}$, from which $\omega$ predicts the next observation $\hat{\mathbf{o}}_{t+1}^{(b)}$.

The optimization problem to simultaneously train $\varphi$ and $\omega$, is the minimization of the following objective function (based on the next observation prediction error of **Eq. 1**):

$$\mathcal{L}\left(\boldsymbol{\theta}_\varphi, \boldsymbol{\theta}_\omega\right) = \frac{1}{B}\sum_{b=1}^{B} \left\|\omega\left(\mathbf{s}_{t+1}^{(b)}\right) - \mathbf{o}_{t+1}^{(b)}\right\|_2^2 \tag{10}$$

We compute this objective function after all $B$ agents have executed their actions $\mathbf{a}_t^\pi(b)$, and let the backpropagation compute the partial derivatives of this objective function with respect to the parameter sets $\boldsymbol{\theta}_\varphi$ and $\boldsymbol{\theta}_\omega$. One gradient descent on this loss is what we call a training step on $\mathcal{L}(\boldsymbol{\theta}_\varphi, \boldsymbol{\theta}_\omega)$.

### 3.3.1 Update Interval

The inverse model and the two discovery policies are trained in parallel to the above training. For losses other than $\mathcal{L}(\boldsymbol{\theta}_\varphi, \boldsymbol{\theta}_\omega)$, instead of performing a training step after every agent executed its action, it is performed after a chosen update interval ($T_\pi$). Since the policy optimization is much more sensible to the i.i.d. hypothesis (of the Robbins-Monro's conditions for stochastic gradient descent to converge (Robbins and Monro, 1951)), we use the largest possible sampling period $k$ for these two types of optimization ($k$ also corresponds to the number of training steps whose the parameters of $\varphi'$ are delayed). To do this, we specify an update interval $T_\pi \in \mathbb{Z}^+$ defining the number of time steps before a training step is performed on the parameters of the inverse model and of the two discovery policies. Given a chosen batch size $B_\pi \in \mathbb{Z}^+$ and the number $B$ of agents running in parallel, a batch of training examples is formed of $\lfloor \frac{B_\pi}{B} \rfloor$ samplings. To maximize the independence between each of these samplings, we choose the sampling period to be $k = \lfloor \frac{T_\pi B}{B_\pi} \rfloor$.

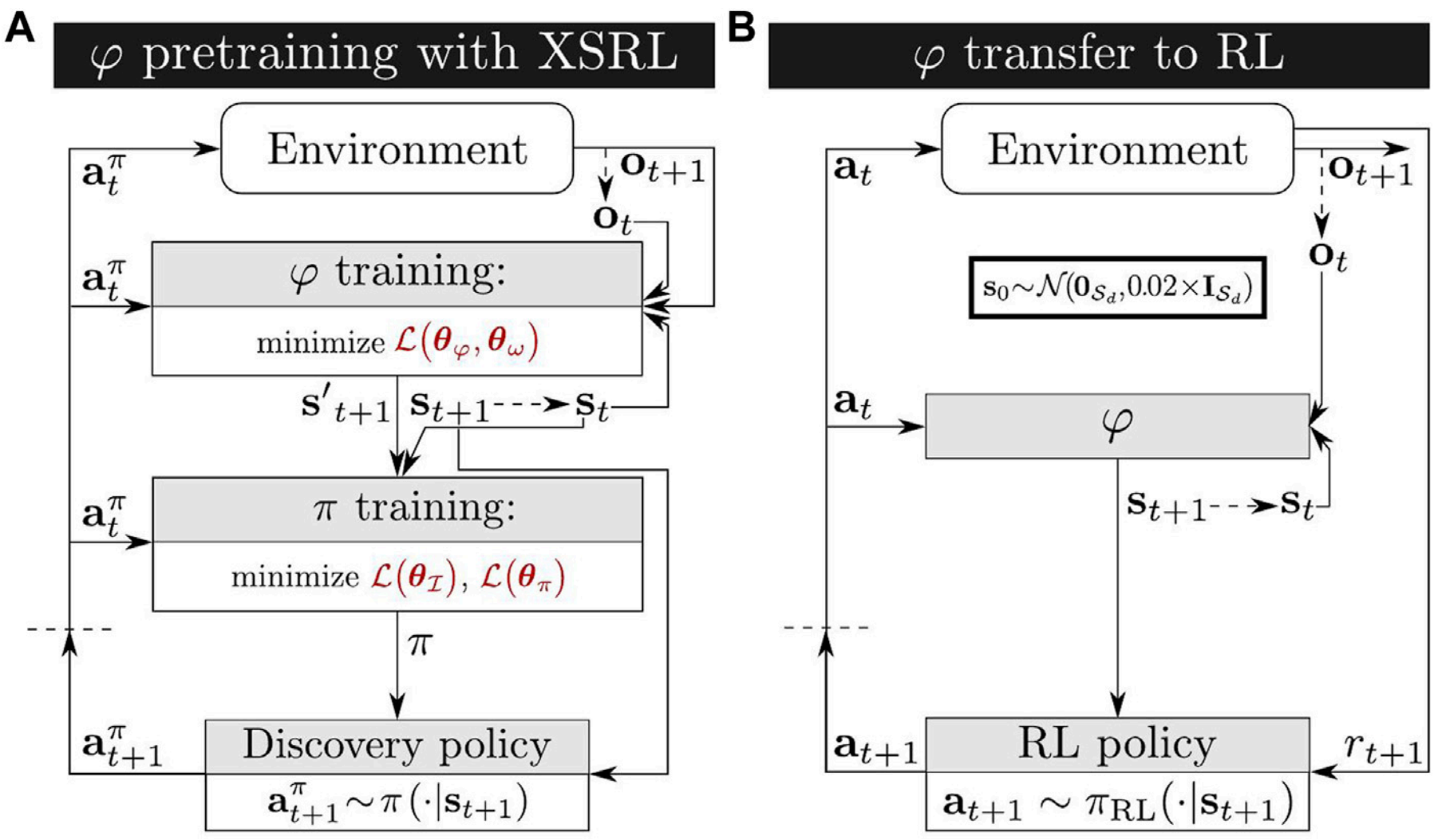

**FIGURE 2 | (A)** Schematic representation of the XSRL twofold training procedure to provide compact state representations by jointly training a state transition estimator $\varphi$ with a next observation predictor $\omega$, guided by two discovery policies $\pi \in \{\pi_1, \pi_2\}$ in an online manner (see **Section 3.3**). Here $\mathbf{s}_{t+1} = \varphi([\mathbf{o}_t, \mathbf{s}_t, \mathbf{a}_t])$, and $\mathbf{s}'_{t+1} = \varphi'([\mathbf{o}_t, \mathbf{s}_t, \mathbf{a}_t])$ where $\varphi'$ is a clone of $\varphi$ whose parameters are delayed by $k$ iterations and kept frozen. **(B)** A schematic illustration of the transfer of the pretrained state representation model ($\varphi$) to an unknown RL task.

The optimization problem to train the inverse model is the minimization of the following objective function (based on the action prediction error of **Eq. 4**):

$$\mathcal{L}(\boldsymbol{\theta}_\mathcal{I}) = \frac{1}{B_\pi} \sum_{i=0}^{\lfloor \frac{B_\pi}{B} \rfloor - 1} \sum_{b=1}^{B} \left\| \mathcal{I}\left(\mathbf{s}_{t+1-ki}^{(b)}, \mathbf{s}_{t-ki}^{(b)}\right) - \mathbf{a}_{t-ki}^{\pi}(\mathrm{b}) \right\|_2^2 \quad (11)$$

The backpropagation computes the partial derivatives of this objective function with respect only to the parameter set $\boldsymbol{\theta}_\mathcal{I}$.

The optimization problem to train the two discovery policies is the minimization of the following objective function (based on the loss of **Eq. 9**):

$$\mathcal{L}(\boldsymbol{\theta}_\pi) = \frac{1}{B_\pi} \sum_{i=0}^{\lfloor \frac{B_\pi}{B} \rfloor - 1} \sum_{b=1}^{B} -\left[ w_\mathcal{I} r^\mathcal{I}\left(\hat{\mathbf{a}}_{t-ki}^{\pi(b)}, \mathbf{a}_{t-ki}^{\pi}(\mathrm{b})\right) + w_{\mathrm{LPB}} r^{\mathrm{LPB}} \left(\mathbf{o}_{t-ki}^{(b)}, \mathbf{s}_{t-ki}^{(b)}, \mathbf{a}_{t-ki}^{\pi}(\mathrm{b})\right) - w_\mathcal{H} \log\left(\mathbf{a}_{t-ki}^{\pi}(\mathrm{b})\right) \right] \quad (12)$$

where the parameter set $\boldsymbol{\theta}_\mathcal{I}$ is frozen, and that of $\varphi'$ is updated every $k$ iterations with that of $\varphi$ and kept frozen. More specifically, the backpropagation computes the partial derivatives of this objective function with respect to the parameter set of $\pi \in \{\pi_1, \pi_2\}$. This objective function is low where the inverse model fails to predict actions, and the predictions of the forward model ($\varphi$) vary greatly.

Finally, to automatically tune the temperature $w_\mathcal{H}$, we minimize the following objective function:

$$\mathcal{L}(w_\mathcal{H}) = \frac{1}{B_\pi} \sum_{i=0}^{\lfloor \frac{B_\pi}{B} \rfloor - 1} \sum_{b=1}^{B} w_\mathcal{H} \left[ -\log\left(\mathbf{a}_{t-ki}^{\pi}(\mathrm{b})\right) - \bar{\mathcal{H}} \right] \quad (13)$$

As explained in **Section 3.2.1**, we choose to simultaneously train two discovery policies to mitigate the "over-commitment" (Shyam et al., 2019). Specifically, our XSRL algorithm (as displayed in Algorithm 1) resets the policy with the lowest accumulation of the two main intrinsic reward terms, which are the prediction error of the inverse model (**Eq. 4**) and the $k$-step learning progress bonus (**Eq. 5**). This accumulation is computed by summing over the indices ($b$) and $i$ as in **Eq. 12** and also over $T_{\mathrm{reset}}$ time steps (defined in **Table 3**), which results in:

$$\sum_{i=0}^{(\lfloor \frac{B_\pi}{B} \rfloor - 1) * \lfloor \frac{T_{\mathrm{reset}}}{T_\pi} \rfloor} \sum_{b=1}^{B} w_\mathcal{I} r^\mathcal{I}\left(\hat{\mathbf{a}}_{t-ki}^{\pi(b)}, \mathbf{a}_{t-ki}^{\pi}(\mathrm{b})\right) + w_{\mathrm{LPB}} r^{\mathrm{LPB}} \left(\mathbf{o}_{t-ki}^{(b)}, \mathbf{s}_{t-ki}^{(b)}, \mathbf{a}_{t-ki}^{\pi}(\mathrm{b})\right) \quad (14)$$

where $T_{\mathrm{reset}} > T_\pi$.

**Algorithm 1.** XSRL algorithm

**Initialization:**
An environment **env**.
Randomly initialized networks: $\varphi = \varphi', \omega, \mathcal{I}, \pi_1, \pi_2$.
$B$ agents, half of which are associated with $\pi_1$, and the other half to $\pi_2$.
**while** ($\boldsymbol{\theta}_\varphi, \boldsymbol{\theta}_\omega$) have not converged **do**
Randomly reset the state $\mathrm{s}_0^{(b)}$ for each agent ($\sim \mathcal{N}(0, 0.02^2)$).
Run an episode with each of the $B$ agents in parallel, and at each iteration:
Sample actions for every agent from $\pi \in \{\pi_1, \pi_2\}$:
$$\mathbf{a}_t^{\pi(b)} = \mu_\pi(\mathbf{s}_t^{(b)}) + \boldsymbol{\epsilon}_t^{(b)} \times \Sigma_\pi(\mathbf{s}_t^{(b)}) \quad , \quad \boldsymbol{\epsilon}_t^{(b)} \sim \mathcal{N}\left(\mathbf{0}_{\mathcal{A}_d}, \mathbf{I}_{\mathcal{A}_d}\right)$$
Perform the action with every agent: $\mathbf{o}_{t+1}^{(b)} \leftarrow \mathbf{env}\left(\mathbf{a}_t^{\pi(b)}\right)$
Predict next state representations for all $B$ agents:
$$\mathbf{s}_{t+1}^{(b)} = \gamma\left(\left[\alpha(\mathbf{o}_t^{(b)}), \ \beta\left([\mathbf{s}_t^{(b)}, \mathbf{a}_t^{\pi(b)}]\right)\right]\right)$$
Compute $\mathcal{L}(\boldsymbol{\theta}_\varphi, \boldsymbol{\theta}_\omega)$ from Eq. 10.
Perform a training step on $\mathcal{L}(\boldsymbol{\theta}_\varphi, \boldsymbol{\theta}_\omega)$ w.r.t. $\boldsymbol{\theta}_\varphi$ and $\boldsymbol{\theta}_\omega$.
**every** $T_\pi$ **iterations do**
Compute $\mathcal{L}(\boldsymbol{\theta}_\mathcal{I})$ with Eq. 11 and perfom a training step w.r.t. $\boldsymbol{\theta}_\mathcal{I}$.
Compute $\mathcal{L}(\boldsymbol{\theta}_\pi)$ with Eq. 12 and perfom a training step w.r.t. $\boldsymbol{\theta}_\pi$.
Compute $\mathcal{L}(w_\mathcal{H})$ with Eq. 13 and perform a training step w.r.t. $w_\mathcal{H}$.
Update the parameters of $\varphi'$ with those of $\varphi$ and keep them frozen.
**every** $T_{\mathrm{reset}}$ **iterations do**
Reset the discovery policy with lowest cumulated intrinsic reward (Eq. 14.)
**end while**

In summary, our XSRL algorithm described in Algorithm 1, performs four types of optimization: (i) of a state transition estimator with **Eq. 10**, (ii) of an inverse model with **Eq. 11**, (iii) of two distinct discovery policies with **Eq. 12**, (iv) of an automatic temperature tuning with **Eq. 13**. See **Table 3** for more details on the hyperparameters of our XSRL implementation.

**Figure 2** shows the two phases of XSRL considered in this work. **A**: the twofold training procedure that XSRL follows in order to effectively explore the environment and to estimate state representations consistent with the true state of the system. **B**: the use of the trained representation model $\varphi$ in an unseen RL task.

## 4 EXPERIMENTAL SETUP

This section describes a systematic evaluation of the criteria that the XSRL algorithm should fulfill. XSRL should learn state representations which (i) retrieve information (possibly by memorizing information from past time steps) to garantee that their transitions are Markovian and (ii) filter unnecessary information. Furthermore, XSRL should learn discovery policies which (iii) explore efficiently even in the presence of aleatoric uncertainty. Finally, after the XSRL pretraining, the state transition estimator $\varphi$ must (iv) provide advantageous inputs to solve unseen RL tasks.

We evaluate criterion (i) by measuring the average of the next observation prediction errors on a training dataset and a test dataset. While the former is made up of samples generated during the training process, the latter is carefully designed for each environment, as described in **Section 4.2.6**. Although some parts of the next observations are irreducibly unpredictable, the lower the error, the more likely the transitions are to be Markovian. Furthermore, we compare the observation prediction error of XSRL with the observation reconstruction error obtained by RAE (Regularized Autoencoder (Ghosh et al., 2019)). However, since it is more complicated to predict the next observation from past time step information than to reconstruct it, it is expected that the latter will perform better.

We evaluate criterion (ii) on state representations and criterion (iii) on discovery policies by training XSRL in a TurtleBot Maze environment with artificial aleatoric uncertainty in its transitions. The aleatoric uncertainty is introduced as follows: at every time step, the color of one of the walls, initially in front of the robot, is randomly sampled (see **Figure 5**). Besides, to fulfill the criterion (iii), we perform an exploration evaluation during the state embedding pretraining of XSRL. We measure the average number of training steps on $\mathcal{L}(\boldsymbol{\theta}_\varphi, \boldsymbol{\theta}_\omega)$ before one of the $B$ ($B$ = 32 as detailed in **Table 3**) agents reaches the other end of the maze. Furthermore, since $\|\mathbf{o}_{t+1} - \hat{\mathbf{o}}_{t+1}\|_2^2$ is a useful prediction error measure to quantitatively evaluate the generalization performance of $\omega$ (which is directly related to the performance of discovery policies), a high error measure will indicate that the exploration strategy is not effective. To complete the evaluation of the discovery policy criterion, we also compare two XSRL ablations:

- XSRL-MaxEnt: trains a policy to maximize its entropy estimation by keeping only the entropy term in **Eq. 12**
- XSRL-random: samples actions randomly from the action space.

Here, XSRL-random is expected to give minimal performance, while XSRL-MaxEnt should be worse than XSRL, as it only depends on the policy distribution.

We evaluate criterion (iv) with the transfer of the trained state representation network $\varphi$ to unseen RL tasks. During RL, the environment provides an agent with extrinsic rewards to train an optimal policy, while $\varphi$ transforms large observations into compact state vectors as shown in **Figure 2B**. To rigorously conduct this evaluation, we use a popular RL algorithm with continuous actions—SAC (Soft Actor-Critic) (Haarnoja et al., 2018)—on each of the three environment tasks shown in **Figure 3**. These continuous control tasks (presented in detail in **Section 4.2**) are challenging because of their high-dimensional observation spaces consisting of images. In order to obtain a quantitative evaluation of our results, we compare the performance with other representation strategies detailed below.

## 4.1 Baselines

We compare the performances of XSRL representations on unseen RL tasks to the following five baselines: *ground truth, open-loop, position, RAE, random network*.

Of all these baselines, only RAE (Regularized Autoencoder) (Ghosh et al., 2019) is a state-of-the-art SRL method. We train it using the same three rewardless environments with fixed state initializations as for XSRL (described in **Section 4.2.4**). However, since it has no associated exploration strategy to generate observations, we use either a random policy (which is defined as above for XSRL-random) as previously done by Yarats et al. (2019), or an effective exploration designed with expert knowledge (indicated by the suffix *-explor*). In TurtleBot Maze, this effective exploration corresponds to episodes with 50 time steps, with random actions, and random resets (i.e. random initial states anywhere in the maze). In the two torque-controlled environments, this effective exploration has 0.5 probability to take a random action and otherwise takes an action sampled from an optimal policy pretrained in the RL context (i.e. where extrinsic rewards are available) with SAC from the ground truth state space.

RAE is a deterministic alternative to the variational autoencoder (VAE) (Kingma and Welling, 2014), which preserves the regularizing effect of the latter. To the best of our knowledge, we do not know of any other method than RAE, belonging to the SRL context and that achieves state-of-the-art performance on the torque-controlled tasks of the DeepMind Control Suite (DMControl) benchmark (Tassa et al., 2018) with visual observations (similar to those considered in this article). Specifically, in the DMControl benchmark, Yarats et al. (2019) obtain results in which RAE with the SAC algorithm performs as well as PlaNet (Hafner et al., 2018), a state-of-the-art model-based RL method.

We also use a random network representation in which, instead of training a network (i.e. similar to the $\alpha$ function of XSRL), its parameters are simply fixed to random values sampled from a Gaussian distribution of mean zero and standard deviation 0.02. This strategy without any training was popularized for classification problems by Jarrett et al. (2009) and then for RL tasks by Gaier and Ha (2019).

We use, only in the InvertedPendulum environment, the position baseline which corresponds to position measurements without velocities. The absence of velocities let us show the relevance of such physical dynamic information to solve the swing up task. To achieve a good performance, XSRL must extract this information from the observation of consecutive time steps by memorizing through the recursive loop.

Finally, we use a ground truth baseline, which is a state directly extracted from the environment dynamics (see **Section 4.2** for details in each environment), and an open-loop baseline, where the state is defined as the time step of an agent. Wile the ground truth baseline is expected to constitute an upper bound on RL performance, the open-loop baseline serves as a sanity check. The latter would enable us to validate whether the three RL tasks require closed-loop policies. That is, whether it is necessary to use the agent's perception and proprioceptive information to solve the task, or whether open-loop policy learning strategies may be sufficient. In particular, this gives the minimum performance to beat to show the relevance of different state representation strategies.

We justify the absence of state-of-the-art end-to-end RL baselines such as (Lee et al., 2019; Kostrikov et al., 2020; Laskin et al., 2020; Srinivas et al., 2020), despite their open source implementations, by their too high computational complexity which is impractical in our hardware setting and limited computational time.

## 4.2 Environment Details

We perform our experiments on the three environments presented in **Figure 3** which are all partially observable due to image observations. InvertedPendulum and HalfCheetah belong to the MuJoCo torque-controlled benchmark (Todorov et al., 2012), and we chose their implemention on PyBullet (Coumans and Bai, 2016–2019) for compatibility reasons on our computers.

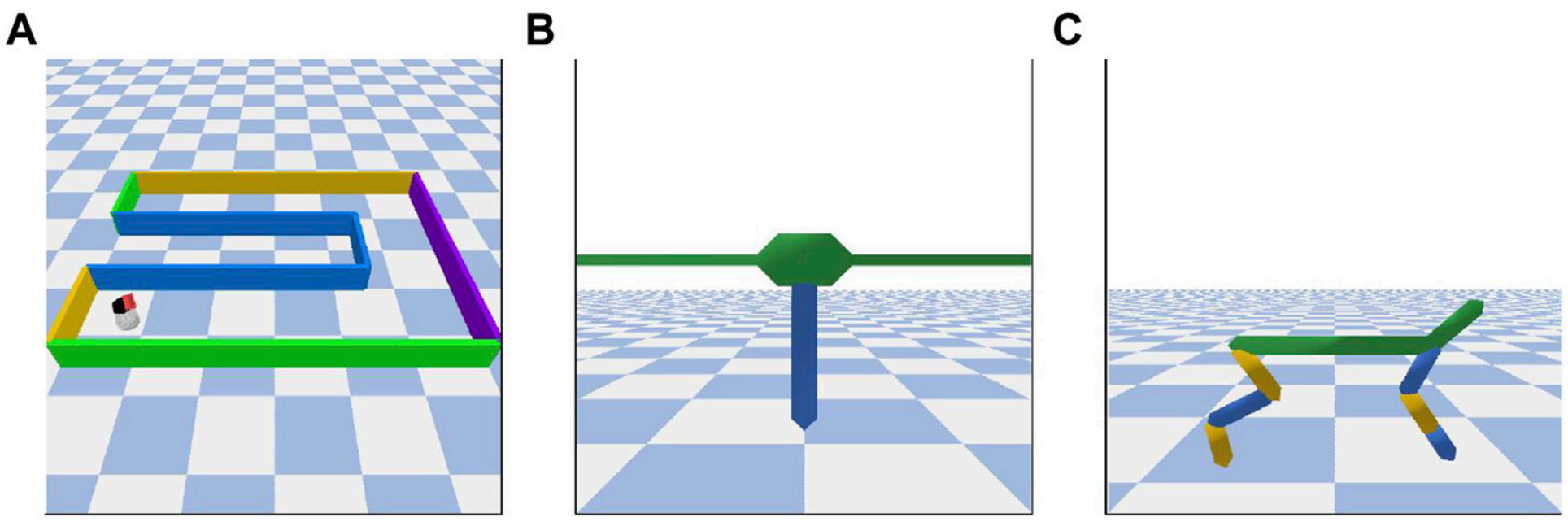

**FIGURE 3 |** High-rendered images of the three continous control environments in PyBullet (Coumans and Bai, 2016–2019). **(A)** The novel TurtleBot Maze environment proposed in this work, where the observation space corresponds to a first-person perspective camera view. We use a goal reaching task in this environment to quantify the exploration performance of XSRL. **(B)** The InvertedPendulum environment provides a swing up task. **(C)** The HalfCheetah environment provides a locomotion task. **(B,C)** are two popular torque-controlled benchmark environments where the observation space corresponds to the view of a camera tracking the agent, as in the DMControl benchmark (Tassa et al., 2018).

#### 4.2.1 TurtleBot Maze

We have implemented this environment as a U-shaped maze with the TurtleBot robot from PyBullet (Coumans and Bai, 2016–2019), inspired by Ant Maze from OpenAI Gym (Brockman et al., 2016) used by Shyam et al. (2019). The two-dimensional action applies a velocity to each of the left and right wheels of the robot. The three-dimensional ground truth state is formed by the cartesian coordinates in $x$ and $y$ axis of the robot and its orientation angle. In this environment, the task consists in a goal reaching task with sparse rewards and a long horizon.[2] Thus, it is a challenge for a RL algorithm to address the exploration/exploitation tradeoff. This task provides a RL algorithm with a sparse reward of +1 each time the robot reaches the goal, a reward of −1 each time it touches a wall, and 0 otherwise, within a maximum of 100 time steps before the robot and the goal are randomly reinitialized. In addition, this task provides a RL algorithm with the position of the goal, which is concatenated to the state representation. Indeed, since the goal position is task-dependent, it cannot be learned by state representations in a reward-free context.

#### 4.2.2 InvertedPendulum

The InvertedPendulum is attached to a pivot point on a cart sliding on a rail. The one-dimensional action applies a force to the cart, which is limited to linear movement on the rail. The five-dimensional ground truth state is formed by the x-axis position and velocity of the cart, the angular position in Cartesian space (i.e. cosine and sine of the angle) and angular velocity of the pendulum. In this environment, the task consists in a swing up task where the pendulum must swing up several times before balancing upward (since the pendulum is initialized downwards). This task provides a RL algorithm with a reward for keeping the pendulum up vertically, within a maximum of 1,000 time steps before the pendulum is reset to a random state.

#### 4.2.3 HalfCheetah

The HalfCheetah is composed of eight rigid links, the torso and the back, and two legs each composed of three rigid and controllable links. The six-dimensional action applies torques to each of the six joints of the two legs. The 17-dimensional ground truth state is formed by the angular positions and velocities of the six joints, as well as agent cartesian position. In this environment, the task consists in a locomotion task where an agent must run to progress as far as possible. This task provides a RL algorithm with a reward for moving the robot as fast as possible, in a maximum of 1,000 time steps and with a constraint that resets it as soon as it gets too close to the ground (which is not applied during XSRL and RAE trainings).

#### 4.2.4 Rewardless Environments

We detail some of the differences in the three environments used without reward in the SRL context and the three tasks described above used in the RL context. In the SRL context (i.e. during XSRL and RAE pretraining), an agent is reset after a longer horizon, and is initialized to a constant state. For TurtleBot Maze the horizon is 500 time steps, hence the need of an effective exploration to reach the other end of the maze, which is at the opposite of the fixed initial state. For the two torque-controlled environments (InvertedPendulum and HalfCheetah), the horizon is 2,000 time steps (so 500 after repeating the action four times). The remaining common hyperparameters of the three environments for the SRL and RL contexts are displayed in **Table 1**.

#### 4.2.5 Image Preprocessing

The image preprocessing performed in these environments follows basically the same state-of-the-art approaches. We divide the pixel values by 255 to normalize them to [0, 1]. Then we downscale the image size to $3 \times 64 \times 64$ pixels just like Mnih et al. (2013); Lillicrap et al. (2015). When the action

[2]In TurtleBot Maze, an agent must perform 47 actions of maximum amplitude to cross the maze.

**TABLE 1 |** Hyperparameters used in the PyBullet environments (Coumans and Bai, 2016–2019).

| Hyperparameter | Value |
|---|---|
| Image rendering size | 3 × 96 × 96 |
| Image size after downscaling | 3 × 64 × 64 |
| Action repeat | 1 TurtleBot Maze<br>4 otherwise |

**TABLE 2 |** Hyperparameters used for SAC [Soft Actor-Critic (Haarnoja et al., 2018)] experiments.

| Hyperparameter | Value |
|---|---|
| Episode length of the environments | 100 TurtleBot Maze<br>1,000 otherwise |
| Discount facor $\gamma$ | 0.99 |
| Replay buffer capacity | 100,000 |
| Optimizer | Adam (Kingma and Ba, 2014) |
| Batch size | 256 |
| Update frequency for the critic target model, and actor model | 2 |
| Learning rate for the critic and actor models, and the automatic temperature tuning | 5e−4 |
| Hidden units of critic/actor models | 128, 512, 128 |

#### 4.2.6 Test Datasets

For quantitative performance evaluation of our XSRL algorithm, we use an error measure of the next observation prediction, and for the state-of-the-art RAE baseline, we use an error measure of the next observation reconstruction. To perform those evaluations, we need an appropriate test dataset for each of the three environments described above. To do this, we carefully collected a wide variety of 400 transitions formed of observation-action pairs into a dataset. We generated them in two different ways. In the case of TurtleBot maze, we hand-designed expert trajectories that follow the U-shape of the maze. In the case of InvertedPendulum and HalfCheetah, we executed a policy learned by SAC from the ground truth state space.

### 4.3 Implementation Details

We now detail the implementation of the training procedures for XSRL and SAC. The source code of our implementation is available online.[3] This implementation uses the deep learning library PyTorch (Paszke et al., 2017). The hyperparameter details for XSRL are detailed in **Table 3**, and for SAC, when different from the original implementation of Haarnoja et al. (2018) in **Table 2**. Preliminary experiments showed that the hyperparameters $w_{\mathcal{I}}$ and $w_{\text{LPB}}$ (to solve the tradeoff during discovery policy training between maximizing the prediction error of an inverse model and maximizing the $k$-step learning progress bonus on $\varphi$) had little impact on final performance.

**TABLE 3 |** Hyperparameters used for XSRL experiments.

| Hyperparameter | Value |
|---|---|
| Episode length for all the environments (after action repeat) | 500 |
| State representation dimension $\mathcal{S}_d$ (i.e. $\gamma$ output dimension) | 20 TurtleBot Maze; InvertedPendulum<br>30 HalfCheetah |
| $\alpha$ output dimension | 30 |
| $\beta$ output dimension | $(\mathcal{S}_d + \mathcal{A}_d)$ |
| Intrinsic reward weight terms | $w_{\mathcal{I}} = 0.5$, $w_{\text{LPB}} = 1$ |
| Optimizer | Adam (Kingma and Ba, 2014) |
| Batch size $B$ for $\alpha$, $\beta$, $\gamma$, $\omega$ | 32 |
| Batch size $B_\pi$ for $\mathcal{I}$ and $\pi$ | 128 |
| Update interval $T_\pi$ for $\mathcal{I}$ and $\pi$ | 512 |
| Reset interval $T_{\text{reset}}$ | 4,096 for both discovery policies |
| Learning rate for $\alpha$, $\beta$, $\gamma$, $\omega$, $\mathcal{I}$, $\pi$, $w_{\mathcal{H}}$ | 1e−4 |
| Hidden units of $\mathcal{I}$, $\pi$, $\gamma$ | 128, 512, 128 |
| Hidden units of $\beta$ | 128, 512, 32 |
| Hidden units of $\alpha$, $\omega$: | |
| $A$ | CNN (strides and filters): (2, 32), (2, 64), (2, 128), (2, 256) MLP hidden units: 1024, 256, 32 |
| $\Omega$ | MLP hidden units: 32, 256, 1024 transposed CNN (strides and filters): (1, 256), (2, 128), (2, 64), (2, 32) |

repeat is one (with TurtleBot Maze), an observation corresponds to the image $\mathbf{o}_t = \mathbf{I}_t$. When it is four (with InvertedPendulum and HalfCheetah), an observation corresponds to the stack of the three consecutive images $\mathbf{o}_t = [\mathbf{I}_{t'-2}, \mathbf{I}_{t'-1}, \mathbf{I}_{t'}]$ of size 9, ×, 64 × 64, just like Lillicrap et al. (2015) and Yarats et al. (2019), where $t'$ corresponds to a time scale four times smaller than that of $t$ (i.e. $t'$ = 4 × $t$). For our XSRL method, this concatenation of images obtained by repeating the last action three times allows not to lose all the information on these time steps. This concatenation of images solves the trade-off between computational complexity and information loss.

For a fair comparison with RAE baseline, the same architecture as $\alpha$ (a convolutional neural network) and $\omega$ (a transposed convolutional neural network) is used for the encoder and decoder respectively. For the RAE and random network baselines, their neural networks similar to $\alpha$ output state representations, while for XSRL the neural network of the forward model ($\varphi$) predicts next state representations. We chose a state representation of 20 dimensions for TurtleBot Maze and InvertedPendulum, and 30 dimensions for HalfCheetah, which correspond to heuristically chosen values. These dimensions were empirically selected to account for the trade-off between sample efficiency and final performance (i.e. between computation time and the optimal policy performance).

[3]https://github.com/astrid-merckling/SRL4RL.

We use the same architecture for the policy (a.k.a. actor model) and the action-value function (a.k.a. critic model) of the SAC algorithm as for the discovery policies, the inverse model and $\gamma$ of our XSRL algorithm. This architecture is made of three-hidden layers (see **Table 3**). The total number of parameters in the corresponding neural network is less than that of the neural network architecture with fewer layers used by Yarats et al. (2019); Hansen et al. (2020) on similar RL tasks, because each layer of our networks is much smaller; see Poggio et al. (2017) for theoretical explanations. As Yarats et al. (2019), we use double Q-learning (Van Hasselt et al., 2015) for the critic model.

The Leaky Rectified Linear Unit (Leaky ReLU) is used for the activation functions between hidden layers, which removes the vanishing gradients encountered with the ReLU and improves the convergence speed and stability (which we observed empirically on preliminary experiments); see Xu et al. (2015) for details.

In our RL experiments, the SAC algorithm is only used to test the generalization of the XSRL state representation to unseen control tasks. This implies that we keep the parameters of $\varphi$ fixed. Due to memory constraints, for all experiments, we use a reduced buffer capacity unlike work comparable to ours: 100,000 instead of 1,000,000 in Yarats et al. (2019).

### 4.3.1 Hardware Details

All our experiments are performed on three computers, each containing 40 cores and a Titan Xp GPU provided by Nvidia.

# 5 EXPERIMENTAL RESULTS

## 5.1 Evaluations of XSRL Representations and Exploration

In this section, we show the results of our quantitative and qualitative evaluations to validate whether XSRL fulfills criteria 1), 2), and 3) which we defined in **Section 4**.

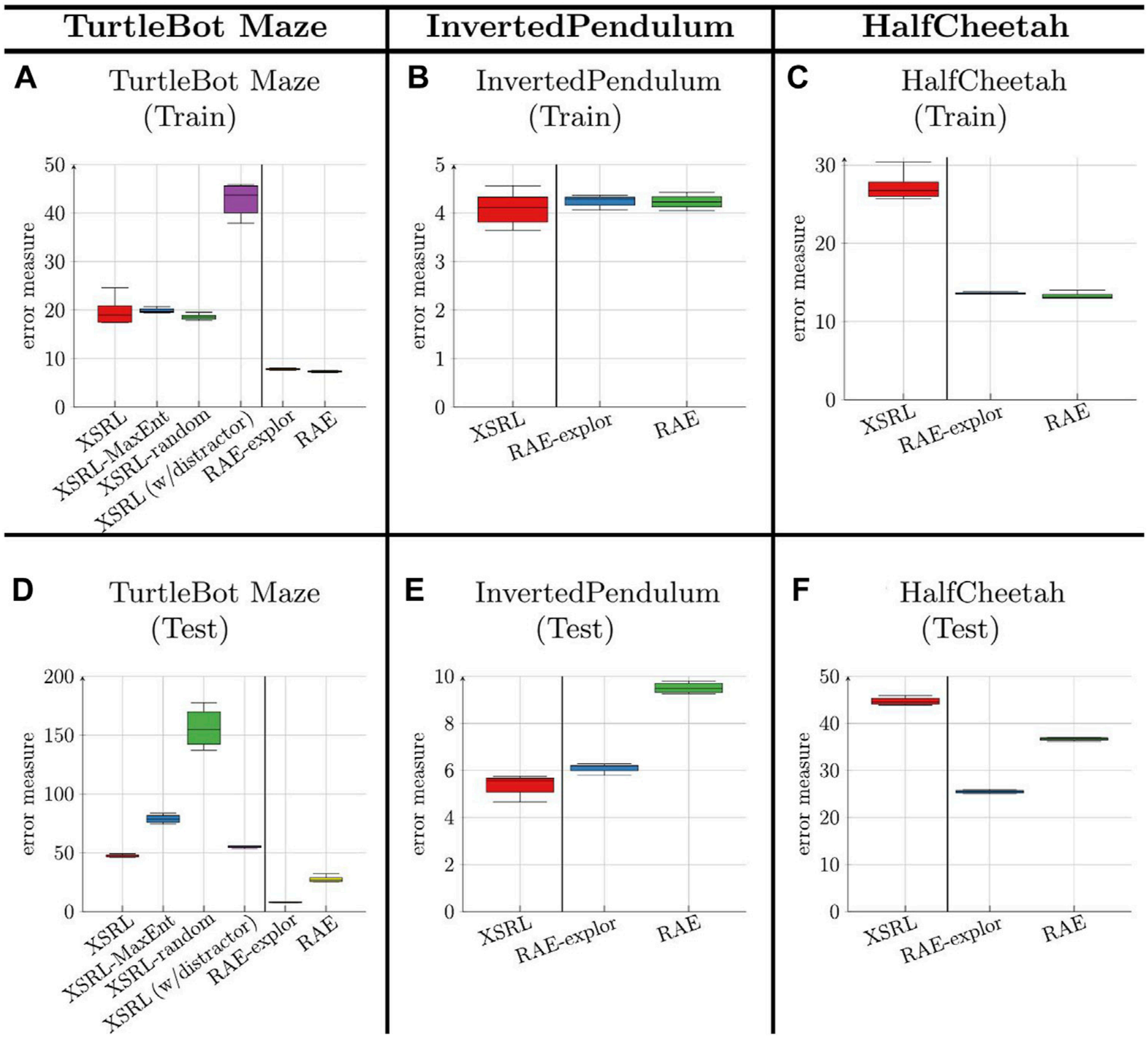

**FIGURE 4 |** Error measure results (the lower the better) obtained on a training dataset (top row) and a test dataset (bottom row) (which is defined for each environment in **Section 4.2.6**), averaged across 5 runs (with different random seeds). This measure corresponds to the prediction of $\mathbf{o}_{t+1}$ with XSRL, and to the reconstruction of $\mathbf{o}_{t+1}$ with RAE. XSRL (w/distractor) is performed in TurtleBot Maze with a randomly sampled wall color (as defined in **Section 4**). XSRL-MaxEnt and XSRL-random are XSRL ablations that follow the entropy maximization strategy and random sampling respectively. RAE-explor benefits from effective exploration (described in \mysec{sec:XSRL_baselines}) while RAE follows only random exploration.

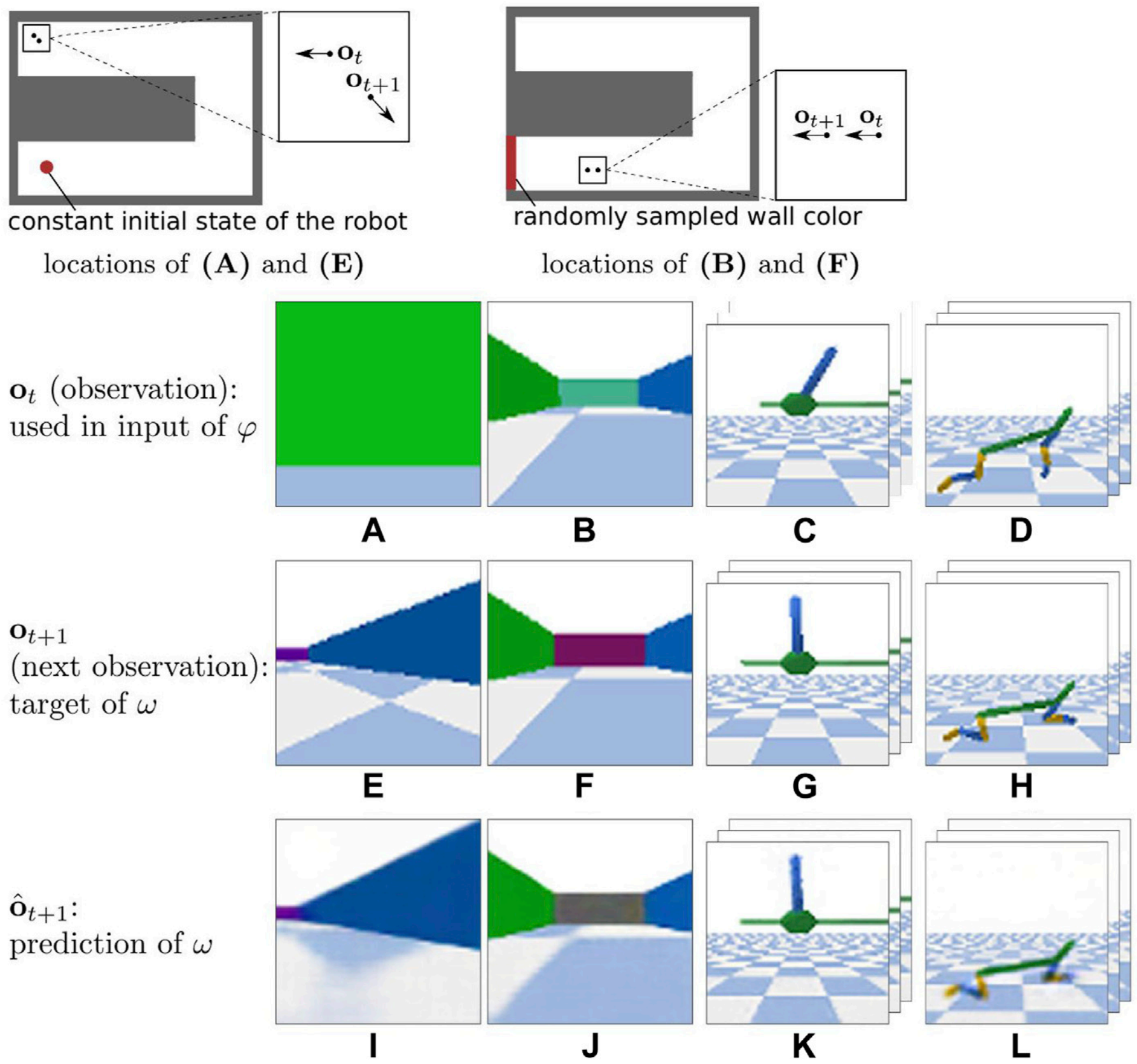

**FIGURE 5 |** One of the most complex transitions in each of the test datasets (defined in **Section 4.2.6**) of three environments. From left to right: TurtleBot Maze, TurtleBot Maze w/distractor, InvertedPendulum and HalfCheetah. The top line shows: (left) the locations of **(A)** and **(E)**; (right) the locations of **(B,F)**. In the bottom line, **(I–L)** show the corresponding $\mathbf{o}_{t+1}$ predictions of XSRL. In **(J)** XSRL tends to filter the random wall color because it predicts a neutral gray color instead. For InvertedPendulum and HalfCheetah environments, as the action is repeated four times, $\mathbf{o}_t$ corresponds to the three consecutive images $[\mathbf{I}_{t'-2}, \mathbf{I}_{t'-1}, \mathbf{I}_{t'}]$ (as defined in **Section 4.2.5**).

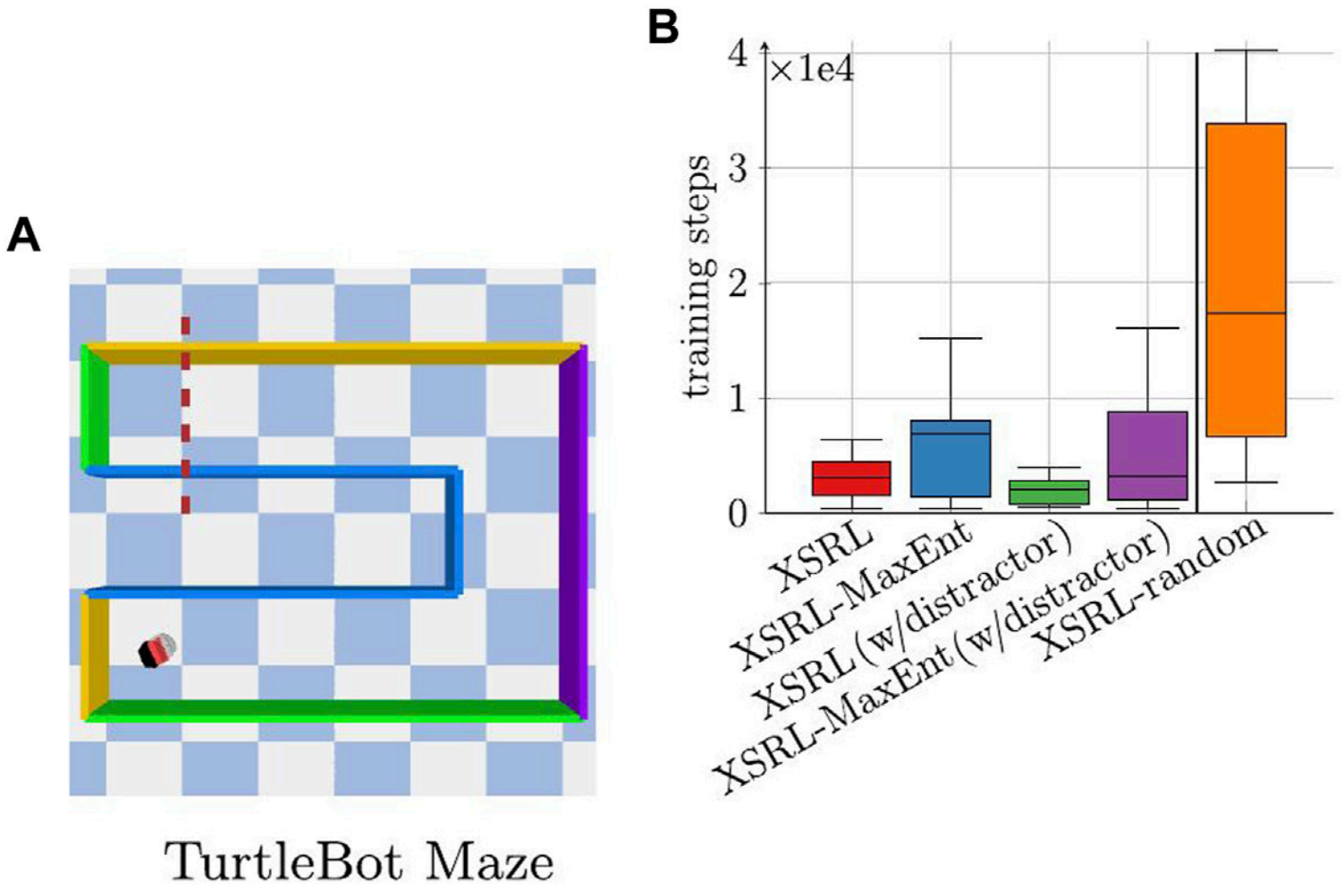

C

| Method | training steps |
|---|---|
| XSRL | $\mathbf{3,261 \pm 1,889}$ |
| XSRL-MaxEnt | $6,593 \pm 5,281$ |
| XSRL (w/ distractor) | $\mathbf{2,458 \pm 1,744}$ |
| XSRL-MaxEnt (w/ distractor) | $6,129 \pm 5,804$ |
| XSRL-random | $26,095 \pm 24,144$ |

**FIGURE 6 | (A)** Shows the top view of TurtleBot Maze where the robot's position corresponds to the constant initial state, hence the complexity of crossing the dotted red line in less than 500 time steps. **(B,C)** Number of training steps on $\mathcal{L}(\theta_\varphi, \theta_\omega)$ before an agent crosses the dotted red line during XSRL training (mean ± standard deviation over 10 runs; the lower the better). Remark: a training step on $\mathcal{L}(\theta_\varphi, \theta_\omega)$ corresponds to a time step for each of the 32 agents. Our XSRL exploration strategy outperforms XSRL-MaxEnt, while XSRL-random provides an upper bound.

**Figure 4** reports the results of the error measure obtained on a training dataset and a test dataset (defined in **Section 4.2.6**) on each of the three environments. This error measure corresponds to the prediction error of the next observation for XSRL and the two ablations (XSRL-random and XSRL-MaxEnt); it corresponds to the reconstruction error of the next observation for RAE (Ghosh et al., 2019) (following a random exploration) and RAE-explor (following an effective exploration) as defined in **Section 4.1**.

We observe on the two environments, TurtleBot Maze and HalfCheetah, that the error measure for XSRL is higher than that for RAE and RAE-explor on both training and test datasets. This does not correspond to a poor exploration performance of XSRL but to the objective function which is more complicated than RAE. Indeed, all information in the next observation that cannot be predicted from the current time step is ignored as it is the case for random distractors or too complex information from the transition model, which tends to increase the prediction error. Furthermore, the qualitative results in **Figure 5** show that XSRL captures well what is relevant to predict the observation that can be explained by agent actions, but ignores less useful/redundant information. For example, in TurtleBot Maze it predicts walls with relatively good precision despite their potentially small size (see the purple wall in **Figure 5I**), but it predicts the checkerboard pattern on the floor in a less accurate way. The former is related to the global information on the topology of the maze, while the latter is not.

The results tend to show that representations learned by XSRL follow Markovian transitions which is criterion (i). Indeed, the representations learned by XSRL can predict the observation change related to robot actions from the current time step only. This is a consequence of the fact that XSRL is based on recursive state estimation predictions (see **Section 3.1**).

In TurtleBot Maze with a distractor represented by a wall color that is randomly sampled after every transition (as defined in **Section 4**), the gray wall predicted by $\omega$ in **Figure 5J** shows that the random colors are ignored by the forward model $\varphi$. Using its forward model, XSRL learns state representations which filter out stochastic information and more generally information that is unnecessary to predict the motion of the system, which is criterion (ii).

We evaluated XSRL discovery policies with a quantitative evaluation of maze exploration presented in **Figure 6**. **Figure 6** shows that XSRL discovery policies lead more quickly to episodes (of 500 time steps) in which agents reach the other end of the maze. Specifically, with XSRL-random, agents can almost never reach the other end of the maze in only 500 time steps. We observe no significant decrease in performance of XSRL with a distractor in TurtleBot Maze. Furthermore, with and without a distractor, XSRL exploration reaches the end of the maze almost twice as fast as with XSRL-MaxEnt. These results tend to confirm that XSRL discovery policies are successful in guiding agents quickly to diverse and learnable transitions, without being affected by the presence of distractors, which is criterion (iii).

The video available here https://youtu.be/IbGa-TC7wek, shows a comparative evaluation between XSRL exploration (left) and random exploration (right) in each of the three environments. It highlights that discovery policies learned by XSRL allow: in TurtleBot Maze to quickly visit transitions far from the initial state position (as shown in the previous results); in InvertedPendulum to balance the pendulum upwards while it is initialized downwards with zero velocity; in HalfCheetah to keep the robot constantly moving and exploring various kinds of postures. We can see that for random exploration: in TurtleBot Maze the robot moves little away from its initial constant state; in InvertedPendulum the pendulum is never upwards; in HalfCheetah it is complicated for the robot to stay in motion since it ends up stuck in a lying position.

In addition to these qualitative and quantitative comparisons, the better performance of XSRL exploration is also confirmed by the quantitative evaluation of the prediction error measure on test datasets for XSRL and its ablations (**Figure 4**). This measure reaches its lowest value with XSRL exploration, followed by XSRL-MaxEnt and finally XSRL-random which is by far the worst strategy.

Apart from the comparative study of our XSRL exploration, we observe that an effective exploration improves the generalization performance of RAE models, which could be expected. Indeed, the quantitative evaluation of the observation reconstruction shows a smaller error on the test dataset with RAE-explor (which is trained with an effective exploration defined in **Section 4.1**) than with RAE (see **Figure 4**).

Qualitative and quantitative performance differences with respect to exploration strategies show the advantage of visiting quickly diverse transitions during state embedding pretraining to obtain better generalization performance over new transitions. However, as we see below, it is only with XSRL that the low error measure translates into good transfer performance with a new RL task.

## 5.2 XSRL Representations Transfer

In this section, we show quantitative evaluations to validate whether state estimators pretrained with XSRL provide advantageous inputs to RL algorithms for solving three unseen control tasks (which is an instance of criterion (iv) defined in **Section 4**). In particular, we use the deep RL algorithm SAC (Soft Actor-Critic) (Haarnoja et al., 2018) which has shown promising results on the standard continuous control tasks InvertedPendulum and HalfCheetah. Throughout these experiments, all parameters of the pretrained state embeddings (with XSRL and RAE) are kept fixed: only the actor and critic neural networks of SAC are trained. We performed 10 runs with different random seeds just like Henderson et al. (2018), Yarats et al. (2019), resulting in 10 different trained policies for each of the representation strategies. For each state embedding pretraining approach (XSRL and RAE) and for the random network, we used 5 different models trained with different random seeds, from which 2 SAC runs with different random seeds are executed. In addition, unlike ground truth, open-loop and position baselines, they transform visual observations into compact state representations of 20 dimensions for TurtleBot Maze and InvertedPendulum, and 30 dimensions for HalfCheetah (as explained in **Section 4.3**).

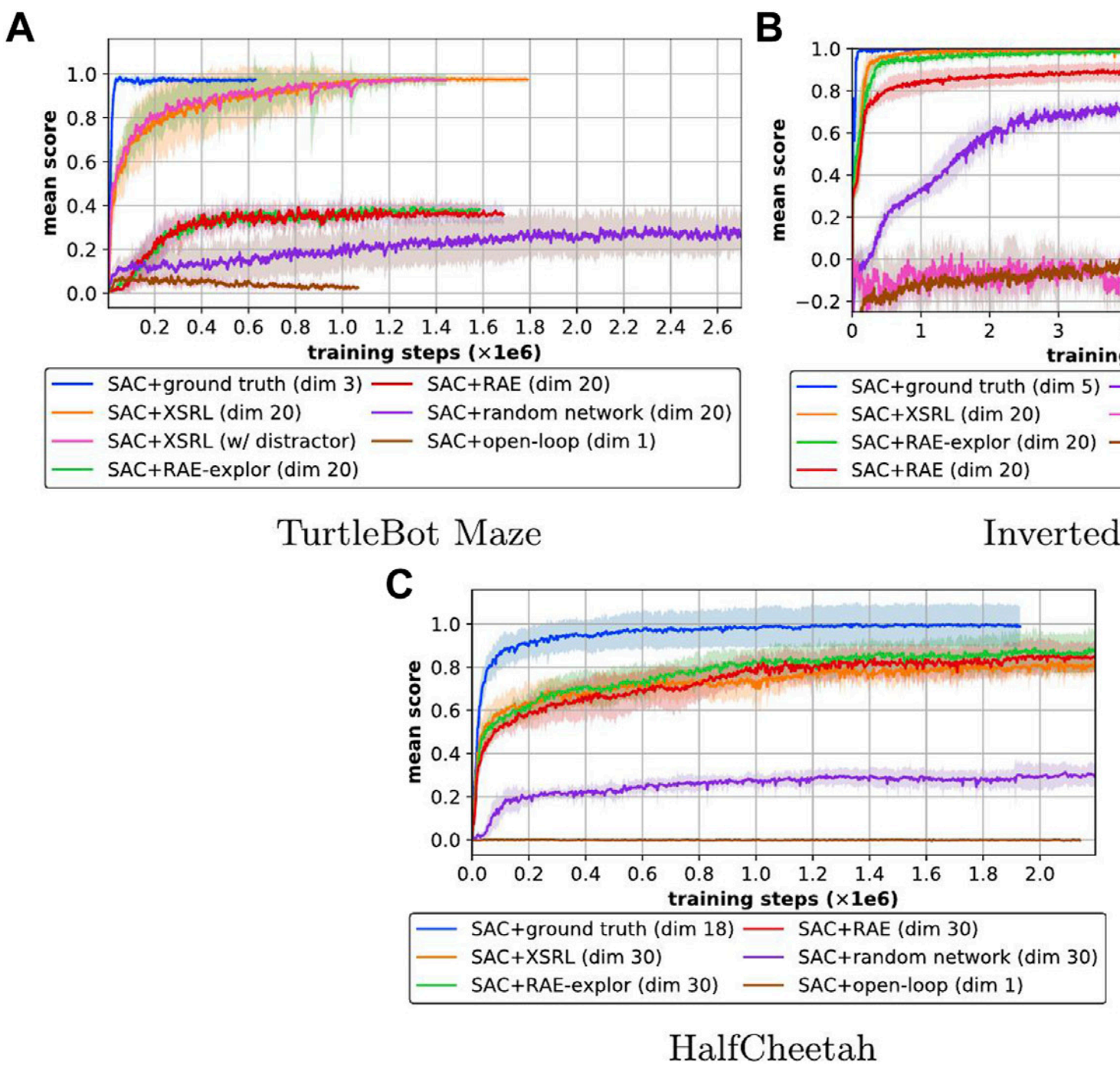

**FIGURE 7 |** Learning curves of the episode returns averaged over 10 episodes (mean in lines and standard deviation in shaded areas over 10 runs; the higher the better) of SAC with different state representation strategies (defined in \mysec{sec:XSRL_baselines}) on three different continuous control tasks. Learning curves of the episode returns averaged over 10 episodes (mean in lines and standard deviation in shaded areas over 10 runs; the higher the better) of SAC with different state representation strategies on three different continuous control tasks. The XSRL pretrained representations are the only one to perform well in three of the environments, while ground truth and open-loop provide an upper and lower bound respectively. A video showing the corresponding learned policies can be found at https://youtu.be/XpRcU75i-iQ.

**TABLE 4 |** Episode returns after convergence of the curves in **Figure 7** averaged over 100 episodes (mean ± standard deviation over 10 runs; the higher the better).

| Mean score | TurtleBot Maze | InvertedPendulum | HalfCheetah |
|---|---|---|---|
| SAC + XSRL | **0.98** ± **0.02** | **1** ± **0** | 0.82 ± 0.03 |
| SAC + RAE-explor | 0.34 ± 0.04 | 0.99 ± 0 | **0.87** ± **0.09** |
| SAC + RAE | 0.34 ± 0.06 | 0.93 ± 0.03 | 0.85 ± 0.08 |
| SAC + random network | 0.27 ± 0.1 | 0.74 ± 0.02 | 0.31 ± 0.05 |
| SAC + ground truth | 0.98 ± 0.02 | 1 ± 0 | 1 ± 0.1 |
| SAC + open-loop | 0.04 ± 0.03 | 0 ± 0.06 | 0 ± 0 |

**Figure 7** shows the learning curves of the episode returns averaged over 10 episodes across 10 different runs. After training, we measured the episode returns averaged over 100 episodes for the 10 different trained policies, which are displayed in **Table 4**. For clarity, we normalized all episode returns between the average SAC + ground truth performance and that of SAC + open-loop, except for the task with TurtleBot Maze as this is evaluated with the probability to reach the goal (from a random initial configuration) in 100 time steps or less. Indeed, SAC + ground truth is an upper bound because it has easy access to the agent's proprioceptive information, and SAC + open-loop is a lower bound because it corresponds to a blind agent. These results show that only XSRL state representations perform well in all three RL tasks, unlike the other state representation baselines.

**Figure 7B** shows that the position baseline does not allow SAC to learn a good policy on the InvertedPendulum task. This confirms that InvertedPendulum and HalfCheetah tasks require information from the positions and velocities of the agent's joints to follow Markovian state transitions which are only related to the local coherence of the environment (Lesort et al., 2018). According to **Figure 7** and **Table 4** on both torque-controlled environments (InvertedPendulum and HalfCheetah) SAC + XSRL and SAC + RAE-explor achieve about the same performance. While on InvertedPendulum they catch up to the ground truth performance, on HalfCheetah they remain slightly below. This is because HalfCheetah is more complex on the control part than InvertedPendulum, as the former has six degrees of freedom and the latter only one.

In TurtleBot Maze, none of the state representation strategies other than XSRL were successful on the navigation task. In addition, a random network is not a viable strategy in any of the three environments, hence the need for a representation learning strategy. Furthermore, as shown by **Figure 7A**, the performance of SAC + XSRL is the same in TurtleBot Maze with a distractor (where XSRL was pretrained with the distractor). This tends to show that XSRL representations can capture information about the environment topology to encode the orientation and position of the robot. As previously explained in **Section 3.1**, this is related to the ability of following Markovian state transitions despite perceptual aliasing (Cadena et al., 2016).

Overall, these quantitative evaluations show that pretrained state estimators with XSRL provide advantageous inputs to solve unseen RL tasks with SAC algorithm, which is an instance of criterion (iv). This confirms that by memorizing the information useful for predicting the consequences of the robot's action in the next observation, XSRL representations can encode the robot's configuration in a state space that exhibits Markovian transitions (useful to control it with RL), while filtering out unnecessary information (useful for generalization on new transitions).

## 6 DISCUSSION

Experimental results show that our proposed XSRL algorithm builds state representations that perform well on three unseen RL tasks. We see the link between the generalization performance of XSRL with respect to its next observation prediction objective (see **Figure 4**) and the transfer performance of its pretrained state estimator ($\varphi$) to a new RL task (see **Table 4**). Specifically, when XSRL achieves good prediction performance on a test dataset, this tends to imply good transfer performance to new RL tasks. On the contrary, our results showed on TurtleBot Maze that the generalization performance of RAE did not guarantee a good transfer to a RL task.

The generalization performance on the test dataset strongly depends on the exploration efficiency (see **Figure 4**), which is better with XSRL than with its two ablations. Our exploration allows agents to reach transitions far away from their initial states and much faster than the policy entropy maximization and random strategies (see **Figure 6**).

Instead of dedicated policies, the exploration strategy could rely on count-based methods (Tang et al., 2017). It might lead to promising extensions of XSRL, with more direct ways to encourage the agent to visit states it has never seen before. However, this approach raises the challenge of keeping the state counts up-to-date and relevant during the whole representation learning process, which requires to constantly update state visitation statistics while the state space changes.

Another promising avenue for XSRL is to extend it to the case where partial observability can be handled not only with memory, but also via active perception (Chrisman, 1992; McCallum, 1993; Whitehead and Lin, 1995). It would both require a modification of the representation learning procedure, in order to take into account information that may be related to hidden aspects of the state, and a modification of the exploration strategy to specifically aim at discovering and exploiting information that removes ambiguity about the true state of the agent.

In this work, we are interested in a state representation that makes the evolution of the system predictable. XSRL tends to filter out information that is unnecessary for this purpose. However, this can be an issue if, in a new RL task, rewards are not related to the evolution of the system. For example, in a task in which an agent must respond to a color signal. Since this information is not controlled by the robot, it will be unpredictable for XSRL and thus filtered from its state representations. Solving this kind of problem is outside the scope of this paper, since we are specifically interested in learning state representations *before* being exposed to various RL tasks and reward signals.

Overall, experimental results have highlighted the main advantage of XSRL in learning state embeddings that can capture both the local coherence of the environment and a global information about its topology. While the state-of-the-art RAE method succeeds in encoding the former, it fails in encoding the latter, and leads to significantly worse results in the TurtleBot Maze navigation task (see **Table 4**).

## 7 CONCLUSION

We have presented a SRL algorithm (XSRL) that trains discovery policies for efficient exploration and pretrains state representations at the same time. Our experiments show that XSRL exploration provides fast maze traversal compared to random policy and policy entropy maximization strategies. Moreover, our comparative evaluation on unseen RL tasks confirms the transfer efficiency of the pretrained XSRL models. One of the most striking results is the superiority of XSRL representations over autoencoder ones, which is due to better representational properties since the constructed states are constrained to follow Markovian transitions. Furthermore, these results highlight the importance of an efficient exploration strategy in state representation pretraining approaches, and more generally in the SRL framework.

## DATA AVAILABILITY STATEMENT

The raw data supporting the conclusion of this article will be made available by the authors, without undue reservation.

## AUTHOR CONTRIBUTIONS

AM: designed the proposed algorithm, implemented the experiments, and also wrote the article. SD, NP-G, and AC: supervised the project and provided guidance and feedback, and also helped with the writing of the article.

## FUNDING

This work has been sponsored by the Labex SMART supported by French state funds managed by the ANR within the Investissements d'Avenir program under references ANR-11-LABX-65 and ANR-18-CE33-0005 HUSKI, and by the project VeriDream that has received funding from the European Union's H2020-EU.1.2.2. research and innovation program under grant agreement No. 951 992.

## ACKNOWLEDGMENTS

We gratefully acknowledge the support of NVIDIA Corporation with the donation of one Titan Xp GPU used for this research.

**Conflict of Interest:** The authors declare that the research was conducted in the absence of any commercial or financial relationships that could be construed as a potential conflict of interest.

**Publisher's Note:** All claims expressed in this article are solely those of the authors and do not necessarily represent those of their affiliated organizations, or those of the publisher, the editors and the reviewers. Any product that may be evaluated in this article, or claim that may be made by its manufacturer, is not guaranteed or endorsed by the publisher.